# An ontology-aided, natural language-based approach for multi-constraint BIM model querying

Mengtian Yin[1], Llewellyn Tang[1,✉], Chris Webster[2], Shen Xu[3], Xiongyi Li[1], Huaquan Ying[4]

[1] Department of Real Estate and Construction, The University of Hong Kong, Hong Kong SAR

[2] Faculty of Architecture, The University of Hong Kong, Hong Kong SAR

[3] Informatics Data Science Function, Royal Berkshire NHS Trust Foundation, UK

[4] Faculty of Civil and Environmental Engineering, Technion-Israel Institute of Technology, Haifa, Israel

u3006144@hku.hk, lcmtang@hku.hk, cwebster@hku.hk, shen.xu@royalberkshire.nhs.uk, landsonl@hku.hk, enochying@campus.technion.ac.il

## Abstract

Being able to efficiently retrieve the required building information is critical for construction project stakeholders to carry out their engineering and management activities. Natural language interface (NLI) systems are emerging as a time and cost-effective way to query Building Information Models (BIMs). However, the existing methods cannot logically combine different constraints to perform fine-grained queries, dampening the usability of natural language (NL)-based BIM queries. This paper presents a novel ontology-aided semantic parser to automatically map natural language queries (NLQs) that contain different attribute and relational constraints into computer-readable codes for querying complex BIM models. First, a modular ontology was developed to represent NL expressions of Industry Foundation Classes (IFC) concepts and relationships, and was then populated with entities from target BIM models to assimilate project-specific information. Hereafter, the ontology-aided semantic parser progressively extracts concepts, relationships, and value restrictions from NLQs to fully identify constraint conditions, resulting in standard SPARQL queries with reasoning rules to successfully retrieve IFC-based BIM models. The approach was evaluated based on 225 NLQs collected from BIM users, with a 91% accuracy rate. Finally, a case study about the design-checking of a real-world residential building demonstrates the practical value of the proposed approach in the construction industry.

## 1. Introduction

Building information modeling/model (BIM) has become a disruptive technology for information management in the global architecture, engineering, and construction (AEC) industry. BIM technology provides a digital building model with semantic descriptions of different types of information [1], which can be utilized throughout the building lifecycle to support various engineering applications, such as design checking [2], energy simulation [3], and facility management [4]. Nowadays, more and more disciplines and stakeholders are incorporated in the context of BIM.

The individual stakeholders manipulate the model from the viewpoints of their varied expertise, and information demands differ noticeably in different processes. Thus, it is crucial to allow BIM users to efficiently retrieve BIM models according to their ad hoc data requirements. On the basis of BIM information extraction (IE) in current practices can be grouped into three categories: schema-based approaches, user interface (UI)-based approaches, and query language-based approaches. Schema-based approaches, such as Model View Definition (MVD) [5], are designed to facilitate data exchange between Industry Foundation Classes (IFC)-based BIM data schema [6] and internal data models of engineering software, based on schematic mapping and transformation. However, the development of model views normally takes long time [5,7], which doesn't meet the ad hoc per-project demands. UI-based approaches allow end-users to select objects in model instances by augmenting interactive UI techniques in BIM tools/software. However, their usability heavily relies on the functionality of software applications. For instance, Autodesk Revit provides filters that only allow filtering class-level elements with attribute restrictions, but object instances and relationships are in vacancy [8]. In contrast, query language-based approaches have greater expressiveness and flexibility in acquiring the ad hoc information subset. There are a few professional query languages specialized for retrieving BIM models, such as BIMQL (Building Information Model Query Language) [9], but the programming principles of these languages form a high entry barrier for common architects, engineers, and project managers [10]. Moreover, it is inevitable but difficult for end-users to comprehend IFC data schema [6] that contains massive and interrelated entities [12,13]. As a result, the above barriers make it difficult and costly for project stakeholders to extract BIM models in building projects.

Taking account of the above problems, a natural language interface (NLI) could be a perspective solution which hides all the formalities of data schema and syntaxes of query languages. By simply submitting queries in natural language, BIM users may be able to obtain the desired model information in a more time and labor-saving manner. Currently, several NL-based BIM model retrieval approaches [11–15] exist, which can process simple natural language queries (NLQs) with fixed patterns (e.g., returning the attribute of an object [13]). However, these methods are not scalable when customized constraints are needed in queries. In practical projects, BIM users may need to query objects based on different combinatory constraints from (a) attributes [10] (e.g., material and properties) and (b) relationships between contextual objects [16]. If multiple-constraint queries cannot be made, the granularity of the NL-based BIM query is too rough to allow the retrieval of much user-specified model information. Herein, multiple-constraint queries represent the queries that contain different forms of constraint conditions (e.g., property, relationship), and these conditions are logically connected if more than one constraint is specified in queries.

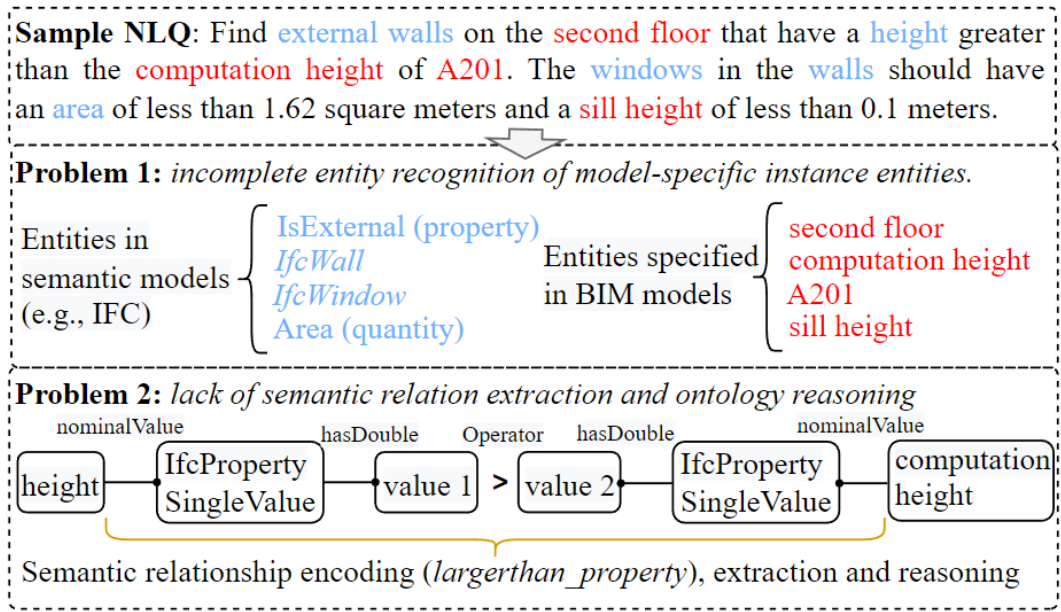

Fig. 1. Problems with processing multiple-constraint queries.

There are several challenges in parsing multi-constraint NLQs against BIM models, which prevents the existing methods [11–15] from outputting the expected results. Consider the example shown in Fig. 1. First, there are lots of model-specific entities (blue entities), which represent object instances and customized properties in BIM models. These entities cannot be recognized based on static semantic models (e.g., ontologies) that are pervasively adopted by the previous studies. Second, there are complex relational constraints between concepts that need to be modeled with reasoning rules and extracted from sentences. Fig.1 provides an example of comparison reasoning between "height" and "computation height", which includes several constructs in the IFC data model and an operator (">"). Unfortunately, the existing methods lack mechanisms to associate semantic relationships with reasoning rules. More than that, the relation extraction in multi-constraint NLQs must consider the dependency and logical connections between concepts, which makes it more challenging.

The above issues reveal the scientific problem of aligning NL texts with IFC BIM models with project-dependent concepts, which makes it impractical to adopt static semantic models for entity recognition and to harness supervised machine learning-based models to parse multi-constraint NLQs due to the lack of training data. To tackle the challenge, this study proposes a novel model-based ontology population and semantic parsing (MOP-SP) approach to automatically convert multi-constraint NLQs into executable codes for retrieving IFC-based BIM models, where semantic parsing (SP) represents transforming natural languages (NLs) into logical forms (e.g., Structured Query Language (SQL)) [17]. The basic idea of MOP-SP is to leverage a seed ontology to incorporate model semantics and utilize natural language processing (NLP) techniques to align NLQs with the ontology for automatically generating IFC-compliant structured queries in an unsupervised manner. A modular IFC Natural Language Expression (INLE) ontology has been developed to supplement the IFC ontologies [18] with the NL expressions of IFC concepts. Then, using ontology-based instance population and text parsing, the intended variables and different levels of constraint conditions in NLQs were extracted and transformed into standard SPARQL queries that can be used to retrieve IFC-based BIM models.

The remainder of this paper is structured as follows. Section 2 introduces the background of this study. Section 3 presents the study's research scope. Section 4 illustrates details of the proposed approach. Section 5 provides a performance evaluation. Section 6 introduces a case study on a realistic building. Section 7 provides a comparison analysis and discusses the limitations. Section 8 concludes by outlining the significance of this research.

## 2. Background

**2.1 Semantic web representation of BIM**

Recently, there has been a research trend to leverage knowledge representation languages (e.g., Resource Description Frameworks (RDF) [19] and Web Ontology Language (OWL) [20]) to enhance the semantic expression and sharing of BIM data [18,21–23]. Numerous research projects have been committed to converting the IFC data specifications [6], an internationally recognized open BIM data standard, into equivalent OWL ontologies [18,21,22,24–26]. Among these, Pauwels and Terkaj [18] present a state-of-the-art (SOTA) conversion pipeline, and their outcomes have been officially adopted by buildingSMART [27]. The resulting ifcOWL ontology is formulated in the OWL2 DL with content that closely matches the original schema, which supports the semantic web functions such as data distribution, querying, reasoning, and reuse of ontologies [18]. Using the available toolkits, the IFC-SPF files (instance models) can be directly converted into schema-compliant RDF graphs.

Because of the complete transformation of EXPRESS constructs, the ifcOWL ontology has critical issues in its incredibly large size and complex instance files. Therefore, Pauwels and Roxin [28] proposed the approach of SimpleBIM that simplifies the ifcOWL graphs and rewrites the data properties. The W3C Linked Building Data (LBD) Community Group [29] developed a lightweight and extensible ontology called the Building Topology Ontology (BOT) that abstracts the topology of buildings, space, stories and building elements [30], which provides a modular approach to selectively incorporate and reuse domain ontologies based on application scenarios [31]. Currently, there are several well-established modular ontologies that address different aspects of BIM, such as Ontology for Managing Geometry (OMG) [32], BuildingElement Ontology (BEO) [33], and the Building Product Ontology (BPO) [34].

### 2.2 Methods for BIM data retrieval and extraction

2.2.1 *Professional programming and query languages for IFC BIM data extraction*

There are a few existing querying approaches for IFC-based BIM data. For example, many application programming interfaces (APIs) are available to ease the use of programming languages to access the model repository. These APIs parse and manipulate the data in EXPRESS format (e.g. JSDAI [35]) or the STEP file of IFC instance data (e.g. Xbim. Essentials [36]).

A step forward was taken by the development of an open query language, BIMQL [9], which processes IFC-based BIM models and allows the filtering, selection, and updating of partial model information. In addition, Daum and Borrmann [37] developed the QL4BIM query language to retrieve spatial and geometrical information in a BIM model using topological operators. Ismail et al. [38] propose the approach of automatically converting IFC EXPRESS schema and IFC-SPF files into IFC Meta Graph (IMG) and IFC Objects Graph (IOG) stored in Neo4j graph database, and using Cypher language to query building data for topology analysis. To lower the entry hurdles for end-users, Preidel and Borrmann [10] introduce a Visual Programming Language (VPL) system named Visual Code Compliance Checking (VCCL), which utilizes user-friendly graphical syntaxes and semantics to represent rules of building codes for compliance checking of digital building models.

As more and more ontologies are emerging for information sharing and integration of BIM, there is a need to effectively query these semantic web data repositories. In current practice, standard query languages like SPARQL (SPARQL Protocol and RDF Query Language ) and SWRL (Semantic Web Rule Language ) are widely used to query ontology-based BIM data. Nikann and Karshenas [39] utilize SPARQL to retrieve the knowledge base from a proposed shared ontology for semantic representation of BIM. de Farias et al. [40] combined the SWRL and ifcOWL ontologies to extract building views. Logic rules should be pre-encoded to specify precise business contexts or processes. To ease the process of semantic retrieval of RDF BIM data, Zhang et al. [16] developed a BIMSPARQL system to retrieve ifcOWL instances based on the SPARQL language. Query functions are extended with respect to common use cases using SPARQL Inference Notation (SPIN).

2.2.2 *Natural language-based BIM data retrieval methods*

Due to the complexity of BIM programming and query languages, several studies proposed to use NLs for BIM model retrieval based on NLIs. The key part to establish NLIs is SP [41], which represents the capability of the system to interpret NLs into logical forms.

Lin et al. [11] propose a NL-based approach for data retrieval of BIM stored in the MongoDB cloud database, with the International Framework for Dictionaries (IFD) [42] leveraged to map concepts in NL to the IFC schema. Nevertheless, their SP method can only process the constraints related to the pre-encoded concepts in IFD, but basic data values (e.g., numeric values) and operators (e.g., ">") are missing for constraining attributes. Wu et al. [48] developed an NL-based retrieval engine that integrates domain ontologies to search BIM object databases and support BIM modeling. However, it cannot be used to retrieve BIM models because it is devised to rank relevant BIM object products on web platforms. Shin and Issa [15] propose a BIMASR (building information modeling automatic speech recognition) framework that uses voice to search and manipulate BIM data, but their SP method only allows to filter walls on a certain building story and change the element properties. Wang et al. [12,13,43] present a spoken dialogue system for speech-based IE from BIM models, which supports several fixed patterns of NLQs, such as returning an attribute value of an object instance. However, different constraints and attribute value restrictions cannot be flexibly assigned in the query. Wang et al. [44] propose a BIM hierarchical tree model (BIH-Tree) for multi-scale information retrieval of BIM models. Although this study addresses multi-constraint queries, the relational constraint solely supports containment relationships that are encoded in the BIH-Tree, and attribute constraint is limited to IFD without filtering operation.

**2.3 Summary**

An NLI sheds new light on allowing practitioners in building projects to query BIM models using human language. However, the current methods have limited capabilities for parsing queries with different kinds and combinations of constraints. The main limitations are categorized into three aspects: First, the attribute constraints cannot be properly extracted because model-specific attribute concepts and different types of attribute value restrictions (e.g., Boolean and literal values) cannot be identified in queries. Second, the relational constraints between contextual objects cannot be retrieved because the semantic relationships (e.g., composition) are not modelled and associated with appropriate reasoning rules. Moreover, the existing methods [11,15] lack a classifier to handle multiple relationships. Third, when there are different logical combinations of constraints in NLQs, the existing solutions cannot extract the logical relationships between these constraints.

To the authors' knowledge, there is no method showing the entire SP process of automatically converting NLs into explicit IFC data schema-compliant queries. Therefore, this study aims to develop the first semantic parser that can translate multiple-constraint NLQs into executable structured queries for NLI applications in BIM-based construction projects. The objectives are to (a) extract complete entities and data value restrictions of attribute constraints in NLQs; (b) extract the relational constraints between contextual objects with their reasoning rules clarified; (c) resolve the logical relationships (LR) between different constraints within one sentence and across multiple sentences to formulate valid queries joining all conditions.

# 3. Research scope

This research focuses on the architectural and structural IFC models because they are the most basic components in building design and construction, including all 19 types of building elements (e.g., *IfcWall, IfcRoof, IfcBeam*, etc.) and 2 types of spatial elements (*IfcSpace, IfcBuildingStorey*).

Following the previous works on BIM query languages [9,16], the supporting constraint conditions for querying BIM models are demonstrated as follows:

- Attribute constraints that allow objects to be filtered by their types (e.g., "select doors with a type of 1250mm x 2010mm"), property (e.g., "find external walls"), quantity (e.g., "walls have a gross volume of 1 cubic meter"), and material (e.g., "walls made of masonry - brick."). The value restriction of property, quantity, and material layer (depth) can be assigned in the queries. Additionally, quantitative comparisons of attribute (property and quantity) values between different objects are supported, including "larger than", "equal to", and "less than" (e.g., "find walls whose width is less than the maximum width of columns").
- Relational constraints between objects, such as containment and composition. This work implements a total of 11 relationships, including 6 commonly used relationships (see Table 1) and their inverse relationships (except space adjacency), for demonstration and testing.

The object mentioned in a query can be either a class of objects (e.g., slab), or a specific object instance in BIM models with its unique name, tag, or long name (e.g., "choose Basic Wall:Exterior - Brick on Block:138157"). Furthermore, the above constraints can be combined with logical operation, including conditional conjunction and disjunction. If a single sentence is not enough to express intentions, our approach supports multiple-sentence queries (queries with more than one sentence).

Table 1. List of relationships and their explanations.

| Relationship | Explanation | NLQ Example |
|---|---|---|
| Containment | A building element is contained in a spatial structure element. | Find walls contained on the first floor. |
| Element composition | An element makes up another element. | Select railings that make up Stair 383509. |
| Space adjacency | A space is next to another space. | Retrieve rooms next to Bath Room. |
| Spatial composition | An object composes a spatial element. | Select spaces on the ground floor. |
| Placement | An object is placed in another object. | Return windows placed in the wall 14341. |
| Space boundary | A building element (e.g., wall) is part of space boundary of a space. | List building elements that enclose room A104. |

The proposed method has two input requirements on the NLQs to ensure the reliability of query results. First, all intended variables (IFC concepts) should be explicitly mentioned in NLQs with their names, and it was assumed that the entity names appearing in NLQs are consistent with the terminology used in target BIM models. The recognition of pronouns and implicit terms is beyond the scope of this study. Second, conjunction and disjunction of constraints on the same side of the target entity must use keywords "and" and "or" to inform the existence of logical connections (e.g., "find walls with a height > 1.5 m and a width < 0.5 m and a top offset = 0 m").

Finally, since this work concentrates on parsing entities and various constraint forms in NLQs, the identification of question types (e.g., "how many") is beyond the scope of the study. The "SELECT" form that returns all variables directly is used as the default query form of the output SPARQL queries.

## 4. The proposed approach

### 4.1 Overview

As illustrated in Fig. 2, the proposed MOP-SP approach comprises four interconnected parts: INLE ontology establishment, model-based ontology population (MOP), ontology-based semantic parsing, and query execution. At start, an IFC Natural Language Expression (INLE) ontology is developed for semantic interpretation of NLQ texts. Then, a MOP is operated to assimilate project-specific concepts that are beyond the already crafted concept realizations in the IFC data model.

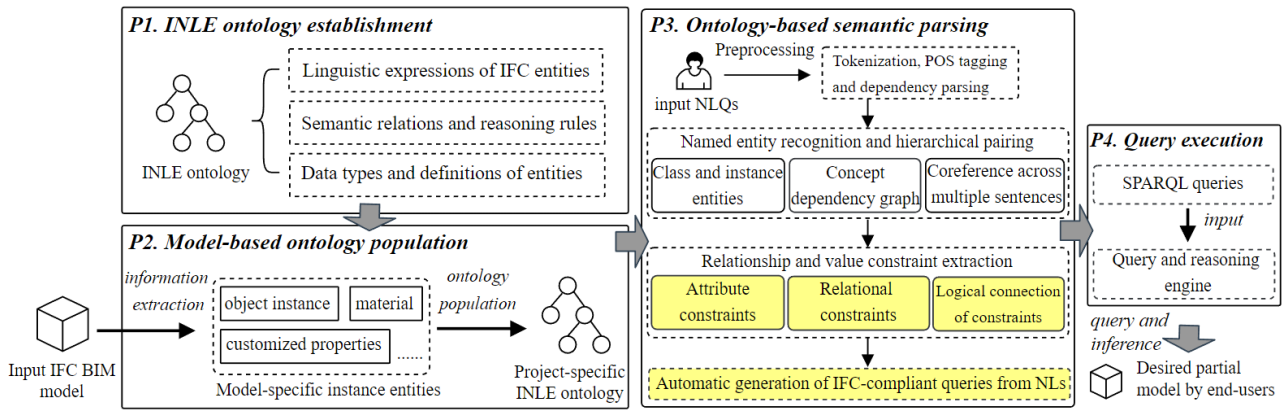

**Fig. 2.** Overview of the proposed MOP-SP method. Contributions are highlighted.

The third part illustrates the working steps of the semantic parser. The named entities (NEs) in the sentence are mapped to concepts in the ontology, and then processed by coreference resolution and hierarchical pairing programs. The relationship and constraint extractions are carried out at three levels: (a) the logical relationships between NEs; (b) the semantic relationships between NEs; and (c) the value restrictions of each single entity. The interpretation results are automatically converted to structured queries by calling the reusable templates. In the last part, the resulting query is executed in an ontological environment to extract the desired partial information subset from BIM models.

## 4.2 INLE ontology establishment

Semantic parsing of BIM queries requires the identification of the mentioned IFC concepts, their relationships, and value restrictions expressed in NLs. Since the current ifcOWL ontology does not have constructs to formally represent the natural language names of IFC concepts, a modular ontology named the IFC Natural Language Expression (INLE) ontology (accessible in Appendix A) was proposed to supplement the ifcOWL ontology with new constructs. The development of the INLE ontology follows the 6-step method proposed by Ding et al. [45]:

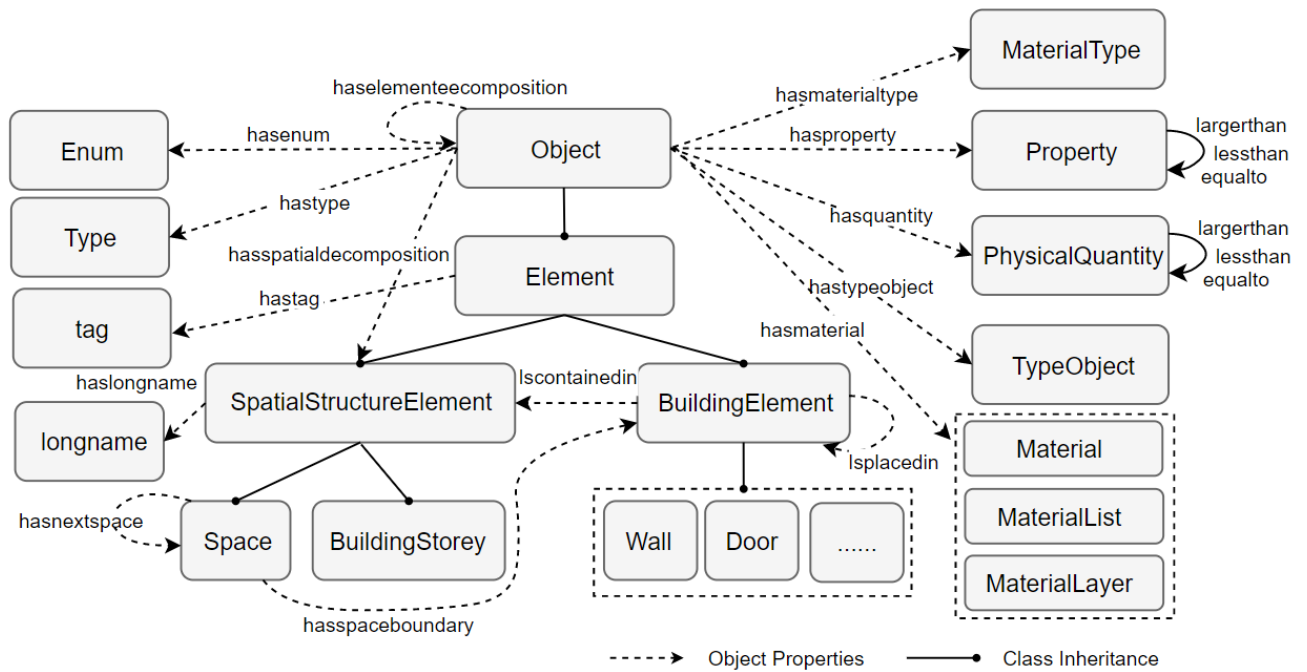

**Fig. 3.** The scope of the INLE ontology for NL-based BIM model retrieval; the prefix of "NLName_ifc" for each class and the suffix "_domain" for each object property are omitted.

- Purpose and scope definition: the INLE ontology was created for semantic interpretation of BIM queries. The scope of implementation complies with the research scope illustrated in Section 3. Only the selected concepts and relationships are modelled in the INLE ontology, as shown in Fig. 3.
- Reuse of existing ontological resources: the ifcOWL ontology [18] were reused to provide ontological concepts for IFC entities, such as the *IfcBuildingElement* and its subclasses. Also, some SPIN rules (e.g., *schm:IsContainedIn*) of BIMSPARQL [16], which define reasoning functions, were employed for query execution.

- Taxonomy construction: a new class called *NLName* (Natural Language Name) was created as the parent class, and its subclasses were associated with the selected IFC entities respectively, following the same class hierarchy of the ifcOWL ontology (see Fig. 3). These classes model the NL names of IFC concepts in two aspects: (a) class name (e.g., "wall" for the entity *IfcWall*); and (b) instance name (e.g., the names of wall instances in BIM models).
- Relation modeling: the semantic relationships between entities are modeled as object properties in the ontology. In OWL, object properties describe binary relationships between individuals. The domain and range were set to be the corresponding *NLName* classes of the IFC entities (e.g., *NLName_IfcWall*), where domain and range are class restrictions of subject and object for a property. These new relationships were assigned annotation properties indicating their definitions and templates for query executions. The template contains the query path and inference rules for ontology-based reasoning over BIM data, which were encoded based on SPIN magic properties. Fig. 4 illustrates an example of an object property *largerthan_property* and its SPIN rule.
- Facets definition: the subclasses of *NLName* are specified with the restriction that the property value of *isnlnamesof* can only be the corresponding IFC classes to show that they are natural language expressions of IFC concepts.
- Instance creation: *NLName* instances were created in the INLE ontology to represent the NL name instances of IFC concepts. For example, the names of the predefined properties in IFC standards were appended in the INLE ontology as instances of *NLName_IfcProperty* with their data types annotated by object property *hasdatatype*. The synonyms, hyponyms, and abbreviations were manually collected from online resources (e.g., WordNet [46]) and added as data property values. In addition, since a name instance might undertake morphological deformations in NLQs (e.g., "IsExternal" *vs.* "External"), different data properties are assigned to name instances to represent various morphological forms, of which the classification is introduced in Section 4.3. Lastly, a data property *hasprojectnaming* was assigned to represent the naming of the concept in the target BIM model, which was set to the original name by default.

In all, there are 121 classes, 58 object properties, 446 individuals, and 3071 axioms that are newly created in the modular INLE ontology. Fig. 5 shows the partial view of the INLE ontology regarding *IfcStair* and its property *HasNonSkidSurface* in the property set definition (PSD). Their relationship was formalized as a new object property *hasproperty_object* in the ontology.

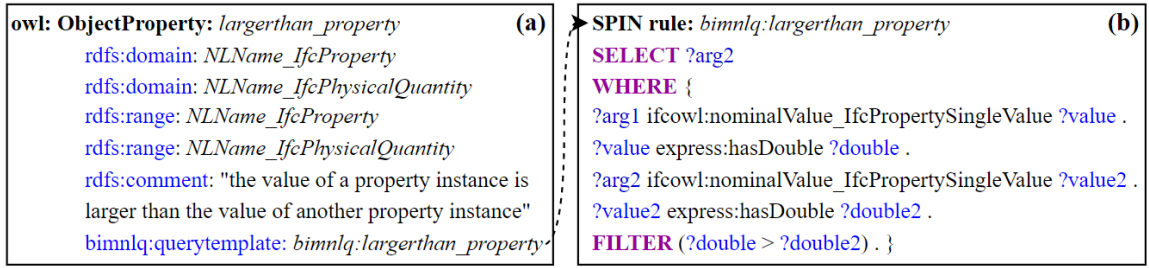

**Fig. 4.** Ontology modeling of semantic relationship *largerthan_property*: (a) object property (b) annotated SPIN inference rule.

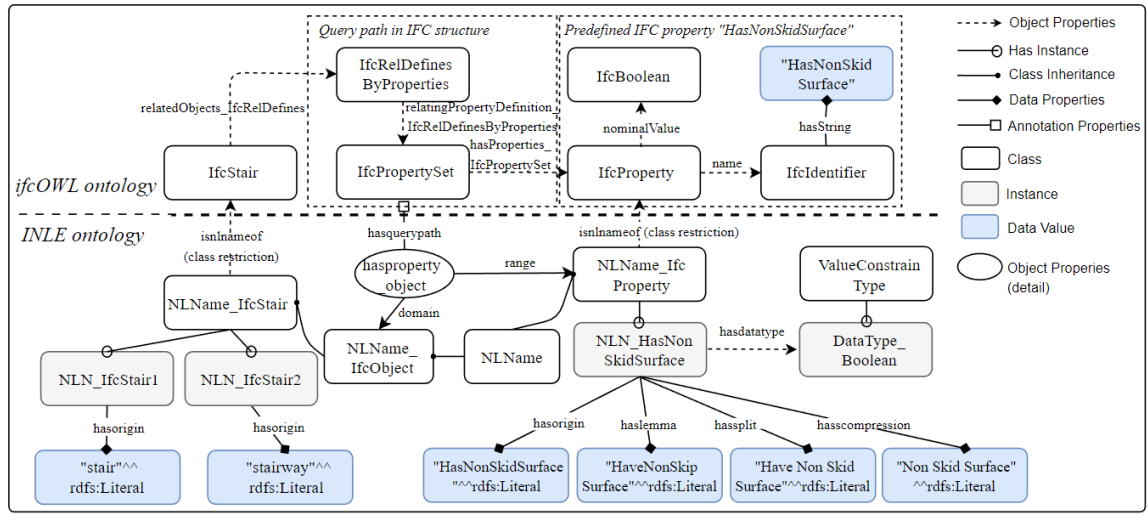

**Fig. 5.** Partial view of the INLE ontology showing *NLName_IfcStair*, *NLName_IfcProperty*, and their relationships. The newly added ontological constructs are presented below the bold dashed line.

## 4.3 Model-based ontology population

AEC projects are complex, project-dependent, and involved multiple stakeholders. A useful NLI should enable practitioners to efficiently retrieve project-specific information contained in BIM models. However, since each project is operated in a unique spatio-temporal context with different designs and engineering solutions, the process and information required to augment BIM models vary from one to another. Thus, there are lots of project-specific information entities that are beyond the scope of the existing BIM ontologies and terminology models (e.g., buildingSMART Data Dictionary (bSDD) [47]). For example, the property "computation height" in BIM models is a customized property that are not covered by the predefined property set definitions (Pset) of IFC data models. Therefore, it is necessary to apply ontology population techniques to absorb project-specific concepts from BIM models before semantic parsing. The ontology population is a subtask of ontology learning [60], the process of appending new instances of concepts or relations into an ontology [61]. In MOP-SP approach, the purpose of ontology population is to extract project-specific name instances from BIM models and populate the INLE ontology with new instances and name variants.

The proposed MOP method is shown in Fig.6. To start with, IE was performed to extract the terms in target BIM models. The range of IE from BIM models is outlined as follows. First, the names of properties, quantities, and materials were extracted from the BIM models to support the retrieval of objects by their attributes. In terms of object type, both *IfcTypeObject* and *objectType_IfcObject* were exploited in IFC (ifcOWL ontology), and thus the names on their labels were also obtained. Second, the name, long name, and identifier (*IfcTag*) of object instances in BIM models were extracted for instance-level retrieval.

Subsequently, the extracted name instances were appended to the INLE ontology under the corresponding ontological classes. For example, "sill height" was added as a fresh instance of *NLName_IfcProperty*. For certain attribute concepts (i.e., property), the data type (e.g., Boolean) is specified for the newly added instances, depending on the types of data values in target BIM models.

The last step is to automatically generate morphological variants of these new instances to obtain the different word forms of concept names. The name strings of the extracted instances and their lemmatized forms were taken as values of the data property *hasorigin* and *haslemma* of ontological instances, respectively. Additionally, since many concept names have orthography conventions with a CamelCase (e.g., "IsExternal"), underscore, or hyphen (e.g., "Masonry-Brick"), these notations were detected in literals and split into separate words to generate the variants for data property *hassplit*. These split names were further pruned when descriptions of concepts in NLQs are incomplete and distorted. Two syntactic patterns prone to undergo deformations were summarized, with rules to generate compression variants illustrated as follows:

- When an entity name is a combination of two tokens with Part-of-Speech (POS) [48] tags of "VBZ" and "ADJ", the tokens with the "VBZ" tag are frequently ignored in NLQs. For example, the property "IsExternal" is often described as "external" in queries. In this condition, the token of "VBZ" was removed, and the remaining segment was taken as a morphological variant for data property *hascompression*.
- The names are in the form of a noun along with a noun modifier connected by a preposition. For example, the property "NumberOfRisers" consists of three tokens: "Number|NN", "Of|IN", and "Riser|NN". In NLQs, the expression often deforms to "Riser Number" with the noun

modifier placed in front and the preposition removed. In response, the token tagged "IN" was dropped, and the remaining tokens switched their orders to form a new compression variant.

Special synonym variants for *IfcBuildingStorey* are automatically generated because the expressions of a building story in NLQs do not always follow the instance names in target BIM models. For example, "Level 1" could be mentioned differently, such as "1<sup>st</sup> floor/level/story", "F1", and "Level One". Hence, additional variants (e.g., "first floor") were formed to cater to diverse expressions of floor levels. Ultimately, due to the variety of terminologies used in AEC projects, the populated ontology probably needs a slight manual adjustment to control the terminology used for NLIs. For example, the new term "computation height" may be regarded as a synonym of the predefined property "height" in some projects, so it should be changed to a synonym variant in the INLE ontology.

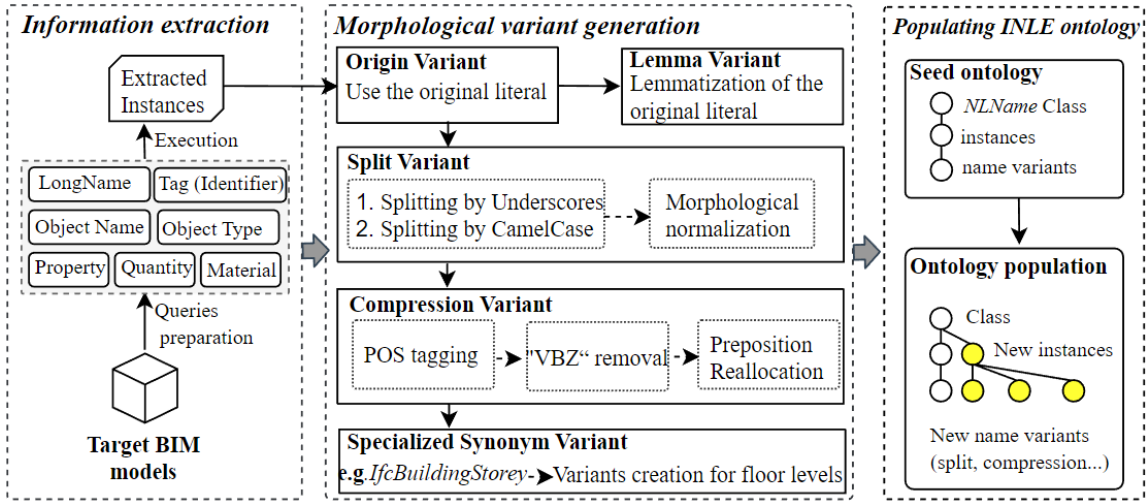

**Fig. 6.** The workflow of the model-based ontology population.

### 4.4 Ontology-based semantic parsing

*4.4.1 Preprocessing*

During the SP process, the NLQs that were typed in were automatically transformed into structured queries. To start with, preprocessing of NLQ was conducted to acquire the grammatical features of the sentences using the existing NLP toolkits. The procedure is illustrated as follows:

(a) Tokenization: transform running texts into tokens [49]. Each word in an NLQ is segmented as an independent token.
(b) Sentence splitting: divide the NLQ text into sentences and index them orderly [50].

(c) POS tagging: POS labels [51] are assigned to each token in NLQ, such as CC (conjunction), VB (verb base form), and NN (noun).

(d) Lemmatization: every word in NLQ is mapped to the standard form.

(e) Dependency parsing: analyze the syntactic structure of a sentence and establishes the relationships between words [49]. A dependency parsing (DP) graph is outputted, with each node representing a token in NLQ and each edge representing the typed dependency from head to dependents. The DP result of a sample NLQ sentence is presented in Fig. 7(a).

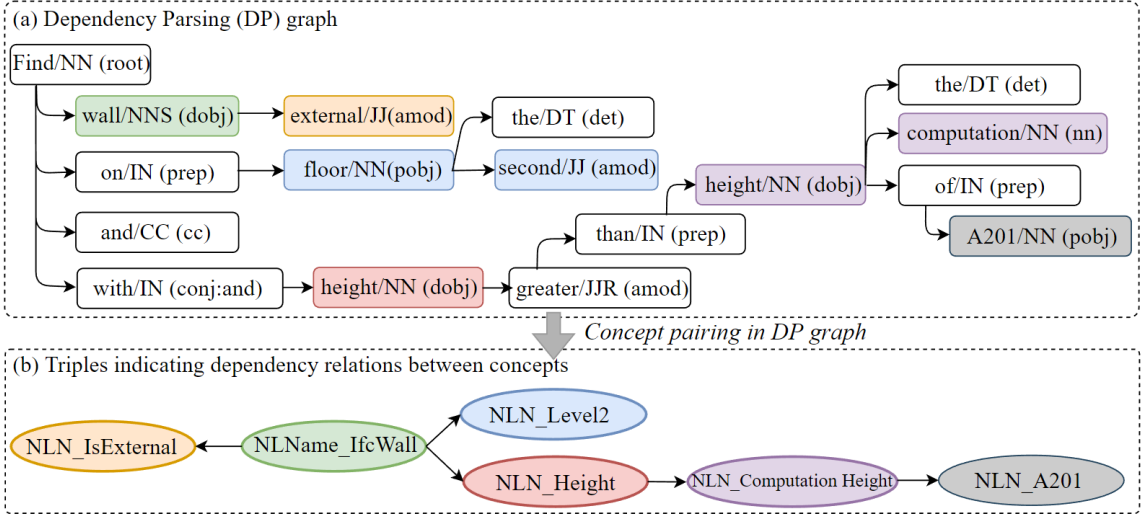

Fig. 7. Dependency parsing, NER, and hierarchical pairing ; (a) dependency parsing graph; (b) concept dependency graph.

4.4.2 *Named entity recognition and hierarchical pairing*

In this section, a named entity recognition (NER) was conducted to match the NEs in NLQ to the ontology concepts. The recognized concepts were disambiguated and were then paired with each other to derive a concept dependency (CD) graph. At the same time, the coreference of entities across different sentences was identified.

(1) Recognition of named entities in NLQs

An OntoGazetteer list was deployed for NER, which contains different names and morphological variants of concepts (e.g., splits, compressions, etc.) associated with concepts' URIs (Uniform Resource Identifier) in the ontology. All names in the list were matched with NLQ sentences based on the Levenshtein Distance [52], which calculates the minimum edit distance between two strings. The threshold for term matching is set to 2 in this study. The information within the identified NEs in an NLQ is stored as a set of {*URI of concept*, *Token Indexes, NE Name, Project Name, Variable*

*Name, Edit Distance, Sentence Index*}. The *URI of concept* is the corresponding URI of the mapped concept in the ontology. *Token Indexes* is a list of indexes of tokens that compose the surface string of NEs. The *NE Name* is the mentioned name in NLQs, and the *Project Name* is the value of the concepts' data property *hasprojectnaming* in the ontology. The *Sentence Index* is the index of the sentences of NLQ where the NE resides. The *Variable Name* denotes the name of variables in the latter SPARQL generation, which is created by concatenating its *Project Name*, the first element in the *Token Indexes*, and *Sentence Index*.

(2) Disambiguation of NER results

The extracted NEs are not guaranteed to be correct because there are false positive (FP) errors that incorrectly recognize the existence of an NE. This results from names that have their constituent tokens matched with other concepts with shorter names. For example, a wall instance called "Basic Wall:Exterior - Brick on Block:143478" will pose a conflicting recognition of "wall" as *NLName_IfcWall*. To address this issue, an element-wise NE disambiguation method was developed by calculating the Identity Score of each NE as follows:

$$Identity\ Score = (1-\partial)\frac{Length(NE\ Name)+1}{Length(NLQ\ sentence)+1} + \partial e^{-(edit\ distance)} \qquad (1)$$

The Identity Score ranges from 0 (negative) to 1 (positive). The first term of the formula awards the name specificity by calculating the proportion between the length of the NE surface string and the length of the input sentence . The second term elicits a greater penalty for NEs with higher edit distance. $\partial$ is a weight factor that controls the influence of both terms, which was set to 0.1 from fine-tuning experiments. For tokens that are part of multiple NEs, the NE with the highest Identity Score is retained, and others are dropped.

(3) Coreference resolution

In some professional use cases, end-users have requests to extract BIM models with complex conditions. Sometimes it is not viable to use a single sentence to precisely express the intention, so multiple sentences are necessary to fully convey the information needs. Comprehension of multiple sentences together entails a coreference resolution (CR) [53] to find all mentions of the same entity in different indexed sentences. For example, the sample query in Fig. 8 could be split into two sentences with a total of 10 NEs recognized. Among these NEs, *wall3_0* and *wall5_1* refer to the same target entity, which should be identified before constructing valid queries.

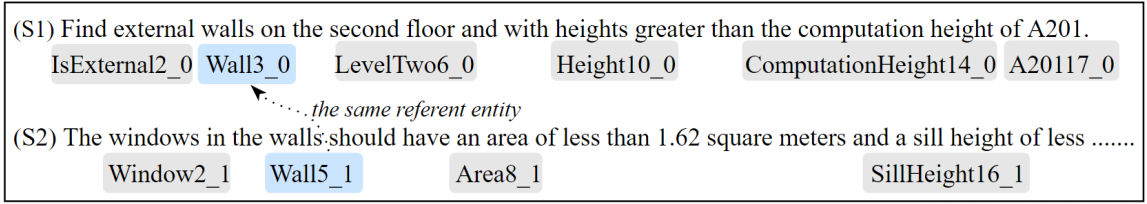

**Fig. 8.** Coreference resolution among multiple sentences in BIM-oriented NLQs. The mentioned NEs are labelled with their variable names.

To realize CR, it was assumed that (a) the same naming is used by end-users across sentences in an NLQ; and (b) the same name mention of objects in multiple-sentence NLQ only refers to one variable. Then, a two-step method was proposed, as shown in Fig. 9. In the first step, the NEs that are not subclasses or individuals of *NLName_IfcObject* are filtered out because the same mentions of attributes (e.g., property, quantity, material) in different sentences probably describe different objects, which is accomplished by acquiring ontological entities of NEs via URI and performing an RDFS (RDF Schema [20]) reasoning to check the class inheritance. In the second step, the values of *NE Name* of NEs in different indexed sentences are compared in pairs. If the mentioned names of the two NEs are identical, they are identified as having the same referent entity. Ultimately, all co-referenced entities are stored for sharing constraints and to generate non-contradictory variables in the final structured queries.

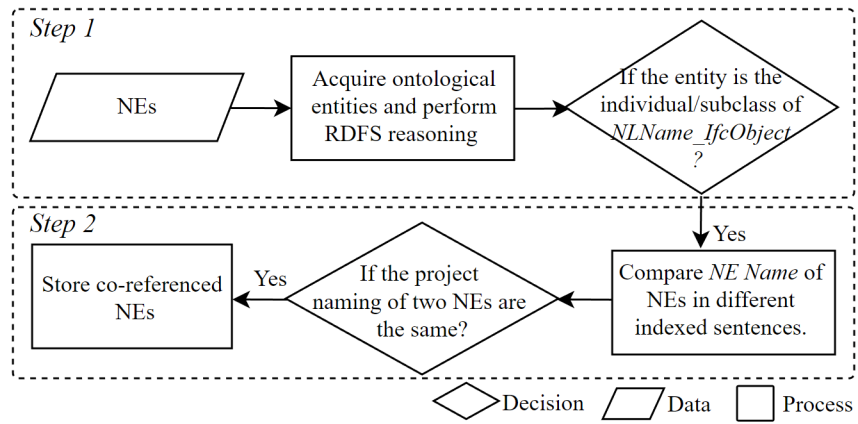

**Fig. 9.** Workflow of coreference resolution.

(4) Pairing NEs in the DP graph

Next, the dependency relationship between the NEs was determined using the DP graph in Section 4.4.1. To make NEs able to be paired in a subtree, the DP graph was first preprocessed by removing edges that pose the cyclic graph structure. Furthermore, because the DP graph takes the predicate of a sentence as the root, the positions of subject and predicate tokens are switched to discover the dependency relationships between concepts as subjects and objects in NLQs.

Having adjusted the heads and dependents of tokens, a recursive analysis was conducted to traverse the DP graph and pair NEs that are nearby in the directed graph. The output, as shown in Fig. 7(b), is a CD graph with a set of edges (triples) in the form of (NE1-Relation-NE2). The first element, NE1, is the head at the upper level in the DP graph. The second element indicates the semantic relationship between NE1 and NE2, which is currently unknown. The third element, NE2, is the dependent beneath NE1 in the DP graph.

4.4.3 *Extraction of relationships and value restrictions*

Until now, we had only been collecting NEs. The next step of SP was to infer the logical forms of NLQs, including logical connections, the semantic relationships between NEs, and the value restrictions of each individual NE.

**Table 2.** Logical relationship forms and examples. "LR" denotes logical relations (Logic-AND and Logic-OR); "SR" denotes semantic relations between objects.

| Logical relation form | NLQ example |
|---|---|
| [Object 1] has [Attribute A] *LR* [Attribute B] | Slab with a length > 5 m and a width < 3 m. |
| [Object 1] has [(*SR1*) Object 2] *LR* [(*SR2*) Object 3] | Windows contained in the 1st floor or placed in the external walls. |
| [Object 1] has [Attribute A] *LR* [(*SR1*) Object 2] | Walls have a top offset > 0 m and bound the space A201. |
| [Object 1] *LR* [Object 2] has [Attribute A] | Columns and walls have a gross volume of less than 10 cubic meters. |
| [Object 1] *LR* [Object 2] has [(*SR1*) Object 3] | Collect railings and stair flights of Stair 866379. |

(1) Detection of logical connections

Two types of logical connections were considered: Logic-AND and inclusive Logic-OR. Logic-AND represents logical conjunctions, and Logic-OR stands for the logical disjunction of conditions. The detection of logical connections was based on keyword searching of tokens "and" and "or" in NLQ sentences. These logical keywords in NLQs could be used in various ways. Table 2 lists five forms that were addressed in this study, where the "Object" refers to building and spatial elements and the "Attribute" represents property, quantity, type, and material constraints. The first three forms focus on the concurrent constraints (attributes or relationships) on the same object, and the last two forms deal with the condition that more than one target object share the same constraints.

As shown in Fig. 10, a graph-based LR extraction method was developed to process the five forms of logical connection and allocate entity triples into logically connected groups. To start with, the nearby NEs on both sides of the keywords "and" and "or" were searched, denoted as L-concept (left) and R-concept (right) respectively. Then, the two NEs (blue and red entities) were checked to see whether they were connected in the CD graph. If so, they were treated as LR-L (logical relation-left) concept and LR-R (logical relation-right) concept. The edge between them was removed, and the parent node of the head NE was searched in the graph. Provided that a parent node exists (scenario 1 in Fig. 10), the LR-L and LR-R concepts indicate two concurrent constraints on the target object, and thus the parent node is additionally linked to the dependent NE; otherwise (scenario 2 in Fig. 10), the LR-L and LR-R concepts represent two logically related target objects that share the same constraints, so their child nodes are shared.

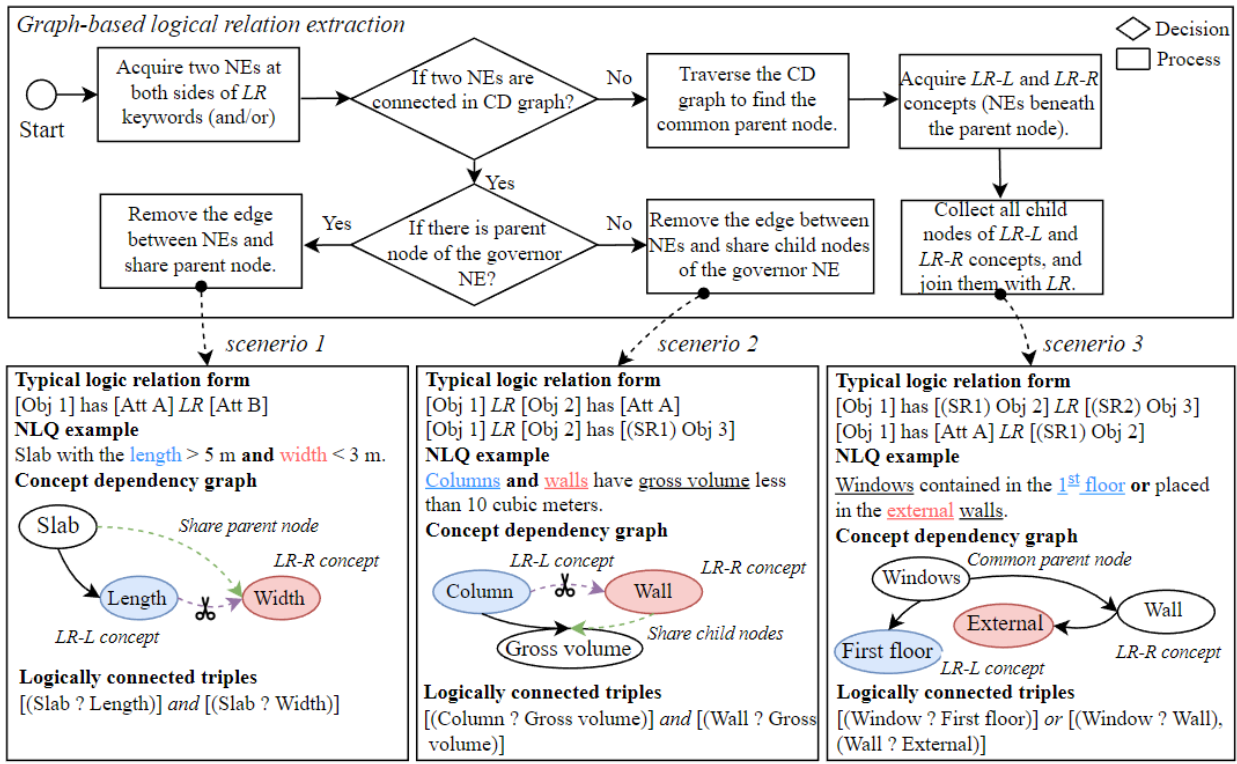

**Fig. 10.** Flow chart showing logical connection detections and how the triples are manipulated.

In the third scenario, L-concept and R-concept are not directly connected. This often occurs in queries that involve semantic relationships with contextual objects, and these objects also have nested constraints. As shown in Fig.10 (c), the spatial entity "First floor" and the property entity "External" are not linked because the latter is used to modify the wall entity. Hence, the strategy is to traverse the CD graph up from the L-concept and R-concept nodes until a common parent node is found. The two child nodes right beneath the parent node along the respective traverse paths are taken as LR-L and LR-R concepts ("First floor" and "Wall" entities in this example).

Lastly, the incoming edges and the subgraphs of both LR-L and LR-R concepts are retrieved in the updated CD graph, therefore obtaining two groups of entity triples that are logically conjunct or disjunct. Note that the first LR form in Table 2 might have complex situations where two attributes are compared with the third attribute (e.g., "stair has riser height and tread depth less than the minimum height of railing"). Hence, two logically connected attributes should also share child nodes if one of them does not have any value restrictions or child nodes in the CD graph. This operation was performed after the value restriction extraction introduced in Section 4.4.4 (3).

(2) Semantic relationship extraction

The remaining triples (edges in the CD graph) have manifestly semantic relationships between resident NEs. A hybrid relationship extraction method was proposed based on ontology and NLP techniques, as shown in Fig. 11. For each triple (NE1–unknown–NE2), the relationships encoded in Section 4.2 were retrieved and an RDFS reasoning was performed to check whether both NE1 and NE2 were instances or subclass/superclass of the object property's domain and range. It is common that there could be more than one relationship that satisfies the class inheritance check. Thus, the candidate relationships were then classified to select the intended one. The classification jointly considers the ontological graph structure and textual information in NLQs. The overall score of the candidate relationships is calculated as follows:

$$OverallScore = (1 - \partial)\frac{1}{GraphDistance+1} + \partial TextSimilarityScore \qquad (2)$$

where $GraphDistance$ measures the depth of inheritance between NEs and domain/range (NE1 versus domain and NE2 versus range) of the candidate relationship [54]. A depth-first graph search algorithm was operated to calculate the shortest distance of direct class subsumption in the OWL graph with the reasoned $SubClassOf$ triples neglected; $TextSimilarityScore$ measures the semantic similarity between text segment in NLQs and the definition of relationships as annotation properties in the ontology, where the text segment used for semantic comparison is the text between head and dependent NEs. Moreover, an unsupervised textual similarity measure [55] model based on WordNet was used in this study. $\partial$ is a weight factor set to 0.1 from our fine-tuning experiment. Finally, the object property with the largest $OverallScore$ was selected, and the triples were updated with the known relationship. Consequently, the relational constraints between contextual objects were all extracted.

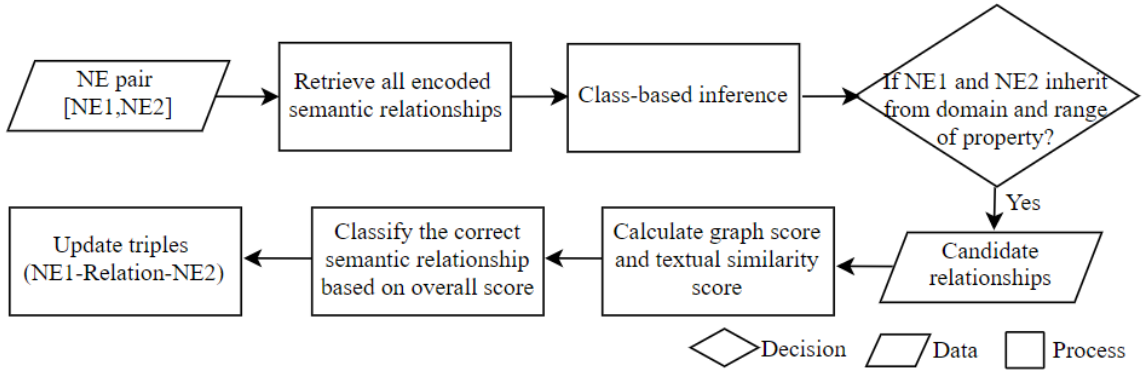

**Fig. 11.** The workflow of semantic relationship extraction.

Fig. 12 illustrates an example of the relational constraint extraction process between the "wall" and the "second floor" in the sample NLQ sentence. The integer above each edge denotes the distance between the NEs and the domain/range, and the float above the object properties denotes the overall score. In this example, *iscontainedin_buildingelement* was correctly classified from four candidate relationships. Based on this result, the inference rule annotated behind this object property could be performed on the incoming BIM data to correctly acquire all instances of building elements contained in the second level.

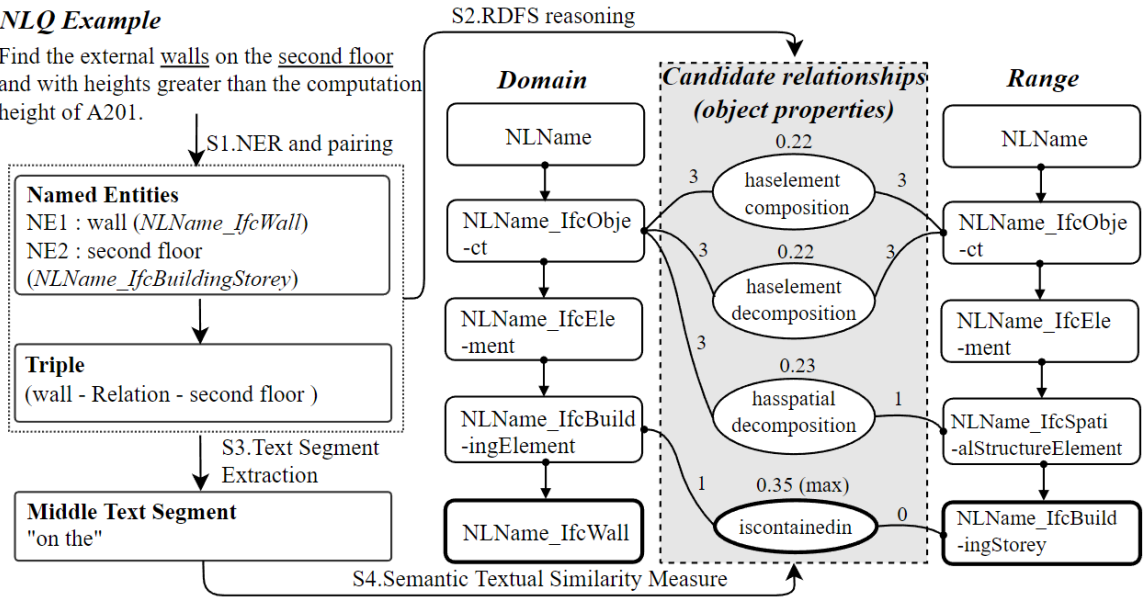

**Fig. 12.** Example of a semantic relationship extraction. The S1, S2, S3, S4 denote the steps in the pipeline. Note that the suffixes of property names indicating domain are omitted in the graph.

(3) Ontology-aided value restriction extraction

In this part, value restrictions on NEs are identified to extract the complete attribute constraints. The attribute concepts to be activated for value extraction include *IfcProperty*, *IfcPhysicalQuantity*, and *IfcMaterialLayer* (i.e., obtain depth restrictions of material layer). The main types of data value restrictions were categorized into quantitative, Boolean, and literal restriction, as explained in Table 3. During the MOP process, the data types of the attribute concepts were identified and encoded in the ontology. Herein, once these concepts were recognized in NLQs, their data types were reasoned in the ontology and the corresponding rules were executed to extract different forms of value restrictions.

To begin with, the text containing constraint information regarding an activated NE was segmented. The text segment included: (a) the nearby token at the left side of the NE; (b) tokens on the subtrees of the NE from the DP graph; (c) tokens of which the indexes are larger than that of the NE and smaller than that of the next occurring NE or logical keyword ("and" or "or") in the sentence. The text segments were treated using different strategies based on the restriction type:

(a) Quantitative restriction: gazetteer lists for comparative (larger than, smaller than, equal to, no larger than, and no smaller than) and maximum/minimum words were compared against the text segment to classify the quantitative restrictions. The cardinal number (CD) in the text segment was then detected if it was a quantitative comparison. For instance, the consequence of attribute constraint extraction of the text "area of less than 1.62 square meters" is (Area–lessthan–1.62).
(b) Boolean restriction: the default value was set to True. Next, negation words such as "not" and "none" were detected, with the times of their occurrence counted. If the count number was odd, then the Boolean value was changed to False. For instance, the restriction extraction of "return the walls that are not load-bearing" would be (LoadBearing–hasbooleanrestriction–False).
(c) Literal restriction: the entire text segment is taken as the restriction value so that the structured query generated in the next section will match the substring of the text segment. For the third example in Table 3, the material of footings named "Concrete–Cast In Situ" is the substring of the extracted text segment "is Concrete–Cast In Situ".

Table 3. Value restriction types and explanations.

| Restriction Type | Explanation | NLQ Example |
| --- | --- | --- |
| Quantitative restriction | The data type is Integer or Double; the restriction is comparative or maximum/minimum. | Slab has the largest area. |
| Boolean restriction | The data type is Boolean; the restriction is to specify True or False. | Walls that are not load-bearing. |
| Literal restriction | The data type is literal; the restriction is to specify the existence of certain strings. | Footings whose material is Concrete–Cast In Situ. |

#### 4.4.4 *Automatic code generation*

Finally, the above SP results were transformed into structured queries for BIM model retrieval. In this study, the RDF-based IFC model (ifcOWL instance) is adopted to represent BIM models due to the increasing popularity of semantic web technologies in AEC industry. Thereafter, the SPARQL query language was employed to retrieve building data.

To automate the code generation process, reusable SPARQL templates were constructed beforehand. According to the query intention, different templates are invoked, and their slots are filled with the results of the previous NLQ interpretation. The outer structure of the SPARQL query was fabricated using the template "SELECT [*variable*] WHERE {[*condition*]} [*sequence modifier*]". The SELECT clause distinguishes the variables in the head of a query. The *Variable Name* of NEs fills in the slot here. The WHERE clause provides a graph pattern to match against the ifcOWL instance. The [*condition*] slot in the WHERE clause consists of three aspects of information:

(a) The identity of variables: for each variable, the corresponding ifcOWL class and the name/longname/tag of the individual should be asserted. The templates in the form of "?[*Variable Name*] rdf:type ifcowl:[*Class Name*]" (assert the OWL class) and "?[*Variable Name*] ifcowl:name_IfcRoot/expr:hasString [*Project Name*]"(assert the individual name) were prepared and filled with the corresponding results.
(b) The relationship between variables: the triples finalized in Section 4.4.3 were operated to produce sentences in the form of "?[*Variable Name1*] [*relationship*] ?[*Variable Name2*]". The template annotated in each predefined object property in the INLE ontology is utilized to clarify the query path and inference between variables (e.g., bimnlq:largerthan_property in Fig.13).
(c) The value restrictions: the extracted restrictions from Section 4.4.3 yield the sentences of "?[*Variable Name*] bimnlq:getvalue ?[*Value*]" initially, then the FILTER clause and built-in operators (e.g., ">", "=", regex) are used to restrict the value variable. A SPIN magic property, *bimnlq:getvalue*, was created to acquire the nominal values of properties and quantities.

If the Logic-OR relationship between NEs exists, the sentences related to LR-L and LR-R concepts were gathered respectively and concatenated using the template "{[*Sentence Collection1*]} UNION

{[*Sentence Collection2*]}". The [*sequence modifier*] slot rearranges the sequence and number of query solutions. When there are quantitative constraints of maximum/minimum in the query, this slot was filled with ORDER and LIMIT modifiers such as "ORDER BY DESC(?[*Value*]) LIMIT 1" (select the maximum). Consequently, a complete SPARQL query was created once all the templates are filled and concatenated. Note that the co-referenced variables from the coreference resolution (see Section 4.4.2(3)) were merged with their constraints combined in logical conjunction. The output SPARQL query of the sample NLQ is shown in Fig. 13. The identity, relationships, reasoning rules, value restrictions and logical connections of variables across two sentences were asserted orderly.

Ultimately, the generated SPARQL queries can be executed to query RDF-based IFC models, therefore obtaining the desired information subsets.

**Input NLQ**: Find external walls on the second floor and with heights greater than the computation height of A201. The windows in the walls should have area of less than 1.62 square meters and sill height of less than 0.1 meter.

**Output SPARQL queries:**

PREFIX ifcowl: <http://ifcowl.openbimstandards.org/IFC2X3_TC1#>
......
SELECT ?external2_0 ?wall3_0 ?A20117_0 ?secondfloor6_0 ?height10_0 ?computationheight14_0 ?window2_1 ?sillheight16_1 ?area8_1
WHERE{
    ?wall3_0 a ifcowl:IfcWall .                                         *Assert identity of variables*
    ?window2_1 a ifcowl:IfcWindow .
    ?secondfloor6_0 a ifcowl:IfcBuildingStorey . ?secondfloor6_0 ifcowl:name_IfcRoot/expr:hasString "Level 2" .
    ......
    ?wall3_0 schm:isContainedIn ?secondfloor6_0 .
    ?wall3_0 schm:hasProperty ?height10_0 .                             *Assert relationships and reasoning*
    ?height10_0 bimnlq:largerthan_property ?computationheight14_0 .     *rules between variables*
    ......
    {
    ?window2_1 schm:hasProperty ?area8_1 .
    ?area8_1 bimnlq:getpropertydoublevalue ?area8_1value .              *Assert value restrictions of variables*
    FILTER ( ?area8_1value<1.62) .
    ......
    }
    UNION    *Union pattern to combine alternative graphs for Logic-OR connections*
    {
    ?window2_1 schm:hasProperty ?sillheight16_1 .
    ?sillheight16_1 bimnlq:getpropertydoublevalue ?sillheight16_1value .
    FILTER ( ?sillheight16_1value<0.1) .
    ......
    }
}

**Fig. 13.** Automatic SPARQL queries generation using cached templates.

# 5. Performance evaluation

## 5.1 Implementation details

All involved programs and algorithms were developed in the Java programming language. Protégé [56] was used to edit the ontology. The RDF and OWL data were manipulated using the APIs of the Apache Jena Framework [57]. For text preprocessing, the Stanford CoreNLP toolkit [50] was adopted. The IFC2×3 TC1 schema [58] was chosen as the ifcOWL ontology.

To demonstrate the usage of the proposed ontology-based semantic parser, a web-based NLI prototype, named NLQ4BIM (Natural Language Query for BIM), was developed with the semantic parsing programs deployed. As shown in Fig. 14(a), its graphical user interface (GUI) was established based on the open-source BIMSPARQL-GUI [16] with a text-based query interface added. Once an NLQ is entered, the automatically generated SPARQL query and query results are presented in the middle-left window and middle-bottom window, respectively. End-users in building projects can directly check the filtered BIM elements graphically shown in the model view window (see Fig 14(b)).

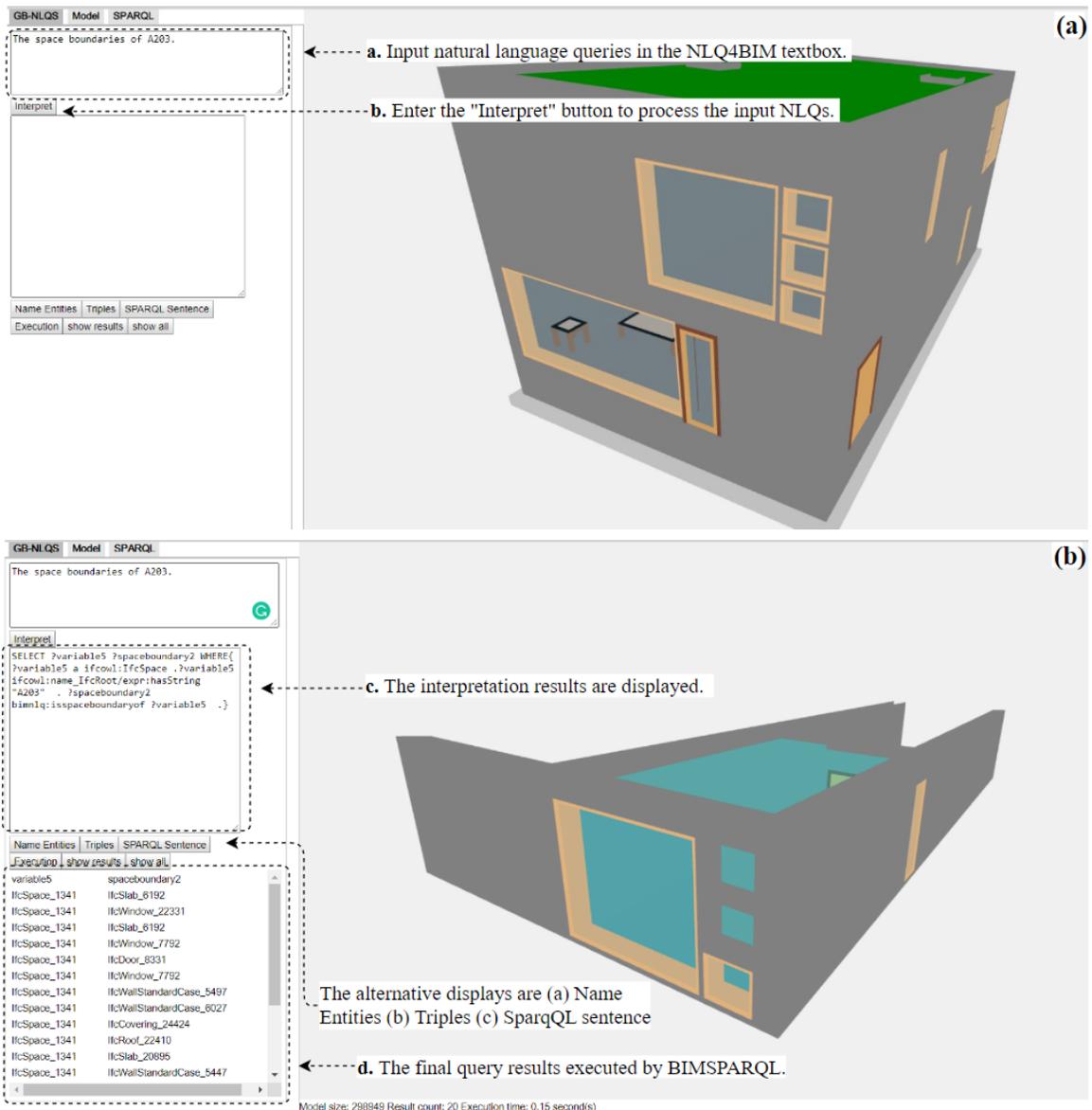

**Fig. 14.** The graphical user interface (GUI) based on BIMSPARQL-GUI [16] to demonstrate the usage of NLQ4BIM (a) and the retrieval results after processing NLQ "The space boundaries of A203" (b).

## 5.2 Experiment design

To evaluate the performance of the proposed approach, ten BIM end-users (7 males and 3 females) were invited to formulate NLQs from the view of their expertise, including BIM model crafters, structural engineers, construction project managers, quantity surveyors, and BIM researchers. They were assigned a sample IFC model (A/B/C) and an information sheet, which demonstrates the scope of the implementation and input requirements. As a result, a total of 225 NLQs, which cover different kinds of constraints listed in Section 3, were collected. The number of variables in individual sentences of NLQs ranges from 1 to 9. Table 4 presents the detailed dataset statistics.

**Table 4.** Descriptive statistics on the NLQ dataset. SD denotes standard deviation.

| Category | Number | Average word count | SD of word count | Average constraint number | Average sentence number |
|---|---|---|---|---|---|
| Single-sentence queries | 188 | 11.68 | 5.88 | 1.87 | 1 |
| Multiple-sentence queries | 37 | 24.4 | 7.60 | 3.76 | 2.16 |
| Total | 225 | 13.77 | 7.78 | 2.18 | 1.19 |

Three sample IFC models are architectural and structural models from open online resources. After approach setups, the accuracy of the query results over 225 NLQs was assessed. The NLQs dataset and the BIM models are accessible in Appendix A.

## 5.3 Test results

### 5.3.1 *Accuracy of the query results*

The accuracy of the SP depended on whether the proposed method could generate a valid query from NLQ, thereby obtaining the correct answers. The resulting structured queries and retrieval results were checked manually against IFC models.

**Table 5.** Accuracy of the overall query results. "Num. of NLQ" denotes the number of NLQs that were tested. "QResult" stands for query result.

| Model | Num. of NLQ | Correct QResult | Accuracy | Accuracy (without MOP) |
|---|---|---|---|---|
| Model A | 75 | 69 | 92% | 8% |
| Model B | 75 | 68 | 90.67% | 18.67% |
| Model C | 75 | 68 | 90.67% | 12% |
| Total | 225 | 205 | 91.11% | 12.89% |

The overall performance of MOP-SP is shown in Table 5. The accuracy rates for Models A, B and C were 92%, 90.67%, and 90.67%, respectively. Hence, the overall performance of the proposed approach was plausible, with a total accuracy of 91.11%. It was found that the proposed approach is capable of handling complex queries that contain multiple variables and constraints. As shown in Table 6, the accuracy rates of extracting attribute constraints and relational constraints were 94.3% and 97.77%, respectively. Table 7 demonstrates five test statements and their processing results. Each query includes more than one constraint with logical connections.

**Table 6.** Accuracy of extraction of different constraints.

| Item | Attribute constraints | Relational constraints |
|---|---|---|
| Total number | 316 | 175 |
| Correct number | 298 | 171 |
| Accuracy | 94.3% | 97.77% |

Another key finding is that the MOP operation has a positive impact on the overall performance. An ablation study was conducted to quantify its effects. As a result, the total accuracy rate of the method without implementing MOP dropped to 12.89% for the three models (see Table 5). This disparity implies that the proposed MOP method significantly improves the overall accuracy because it makes model-specific entities searchable.

Table 7. Examples of the tested queries and parsing results. "N.A." denotes "not applicable".

| Test statements | Result | Sources of errors |
|---|---|---|
| 1. Walls and doors are placed in the level with the highest elevation. The doors should have head heights > 2.0 or trim widths < 0.076. | True | N.A. |
| 2. List non external railings classified as Railing:Garage Security Grille 6'-10" High and having a height higher than the height of railing 1321334 . | True | N.A. |
| 3. Curtain walls belong to Curtain Wall:(existing) Curtainwall - Blank and contains members which are mullions and have a span of more than 17.00. The members are not load bearing. | True | N.A. |
| 4. Walls bound FUEL VAULT and have heights larger than the maximum height of spaces in the second floor . | False | Dependency parsing |
| 5. Select the non-external windows that are located at Level 2 and use metal aluminum as the venetian blind box surface and have a width of 850 mm. | False | Value restriction extraction |

(1) Source of errors

A total of 26 errors occurred, which were categorized into three aspects: (a) NER error; (b) constraint extraction error; and (c) DP error.

Table 8. Precision, recall, and F1-score of the NER.

| Model | Total Num. | Predicted Num. | Correct Num. | Precision | Recall | F1-score |
|---|---|---|---|---|---|---|
| Model A | 265 | 265 | 262 | 98.87% | 98.87% | 98.87% |
| Model B | 278 | 279 | 276 | 98.92% | 99.28% | 99.1% |
| Model C | 273 | 278 | 271 | 97.49% | 99.27% | 98.37% |
| Total | 816 | 822 | 809 | 98.42% | 99.14% | 98.78% |

NER errors accounted for 53.85% of all errors, with its precision, recall, and F1 score presented in Table 8. The errors mainly stem from the name ambiguity problem that implies one name mention in queries can refer to different entities or data values. This led to six FP errors and seven false labeling (FL; the boundary of the surface string is correctly segmented, but the wrong entity type is assigned) errors. The FPs occurred when a concept name overlaps with or matches substrings of a constrained value of another concept, an NE would be mistakenly hypothesized in the sentence. Fig. 15(a) illustrates a test statement that was incorrectly interpreted. The mention of "level 1" should indicate the literal data value of the property "home story", but it was wrongly recognized as an instance of *IfcBuildingStorey*. Similarly, the FL errors occurred because of synonyms of different concepts in the ontology. Fig. 15(b) shows a query statement that causes such a mistake. The mention "roof" can be mapped to different candidate entities due to their identical name descriptions.

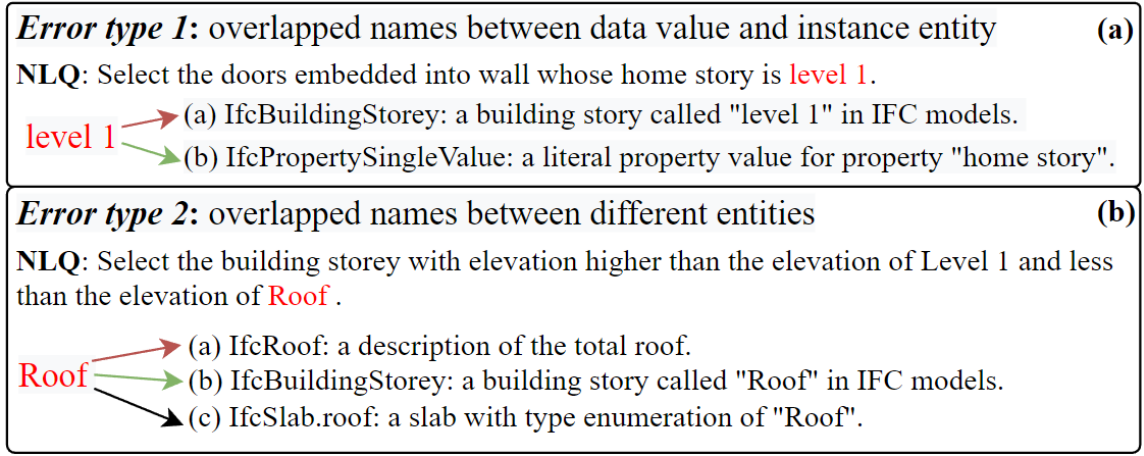

**Fig. 15.** Explanations of sources of errors: (a) FP error; (b) FL error. The red and green arrows represent the prediction and ground truth answers, respectively.

The constraint extraction accounted for 21.43% of all errors. The precision, recall, and F1 score of LR detection, semantic relationship extraction, and value restriction extraction are presented in Table 9, respectively. While LR detection and relation extraction worked perfectly when preceding steps were correctly operated, all errors occurred during the value restriction extraction process because the complete text segments and the intended values cannot be properly extracted. These errors are associated with NER problems, where improper recognition of NEs makes some literal value restrictions impossible to be identified (see Fig. 15(a)). Besides, there are no errors in value extraction for numeric and Boolean data types.

**Table 9.** Precision, recall, and F1-score of the relationship and value restriction extraction. Rel. denotes the relationship, and Res. abbreviates the restrictions.

| Item | Total Num. | Predicted Num. | Correct Num. | Precision | Recall | F1-score |
| --- | --- | --- | --- | --- | --- | --- |
| Logic Rel. | 95 | 95 | 95 | 100% | 100% | 100% |
| Semantic Rel. | 550 | 550 | 550 | 100% | 100% | 100% |
| Value Res. | 178 | 171 | 171 | 96.07% | 96.07% | 96.07% |

Finally, DP induced errors of 25%, which are caused by the preprocessing results of the applied NLP toolkit. In practice, it was found that the dependency parser was more error-prone when dealing with lengthy BIM-oriented NLQ sentences (> 5 variables). The 4[th] query statement in Table 7 shows an example of DP error, where the NE "second floor" was wrongly dependent on the NE

"height". Additionally, the irregular concept names, especially object instance and type names in NLQs, also contribute to DP errors.

5.3.2 *Computation time*

The formalized programs and algorithms were tested on a computer with an Intel ® i7 CPU (2.6 GHz), 16 GB RAM, and the Windows 10 64-bit system. The average time of the data extraction and morphological variant generation for MOP was 170.33 seconds.

Table 10. Average computation time of each step in the experiment.

| Step | Avg. time (s) |
| --- | --- |
| Loading RDF BIM data | 30 |
| NLP preprocessing | 8.67 |
| NER and hierarchical pairing | 4.6 |
| Logical relationship detection | < 1 |
| Semantic relationship extraction | 6.33 |
| Value restriction extraction | < 1 |
| Automatic code generation | 10 |
| SPARQL query execution and inference | 0.6 |
| Total | 60 |

The operation of SP was efficient. The computation time of processing single-sentence queries is shown in Table 10. Loading RDF-based BIM data (the INLE ontology and BIM models) takes up the majority of time (50%). Besides, NLP preprocessing, relation extraction, and SPARQL query generation also cost comparatively long computation time (>5s).

# 6. Case study

To validate the applicability of the proposed MOP-SP approach in processing multi-constraint queries in real-life use cases, a case study was carried out about an application scenario of BIM-based design checking. In the design phase, architects should produce design models that meet a wide variety of functional requirements from owners, stakeholders, and building regulations [59]. Since manually checking models against regulations is time-consuming, tedious, and error-prone, a number of approaches have been proposed to transform building codes into computer-readable rules [60–63] and assess BIM models automatically [2,63–68]. While most existing systems provide rigid functions for rule execution, designers frequently face demands to modify or customize rules based on ad hoc per-project requirements, constraints, and languages. Thus, an NLI would be useful as it enables designers to generate NLQs for design-checking depending on available BIM information.

The case study was conducted with respect to a student residential building at the University of Hong Kong (HKU) [69]. Due to the large number of residents, fire safety is a critical concern in building design for this high-rise building with 29 floors and 389 accommodation rooms (see Fig. 16(a)). To verify whether the proposed approach can be exploited by designers to find objects with design flaws in the building models, the Revit model that complies with the naming convention of the project was collected (see Fig. 16(b) and 16(c)). The NLI introduced in Section 5.1 that deployed the developed semantic parser is used to check the residential BIM model against the fire safety regulation [70] applicable to tall residential buildings in Hong Kong. The 20$^{th}$ section of the regulatory document about the construction of rescue stairways was selected because it contains a set of complex criteria involving logical connections, composition relationships, and quantitative constraints, which comply with the implementation scope of this study.

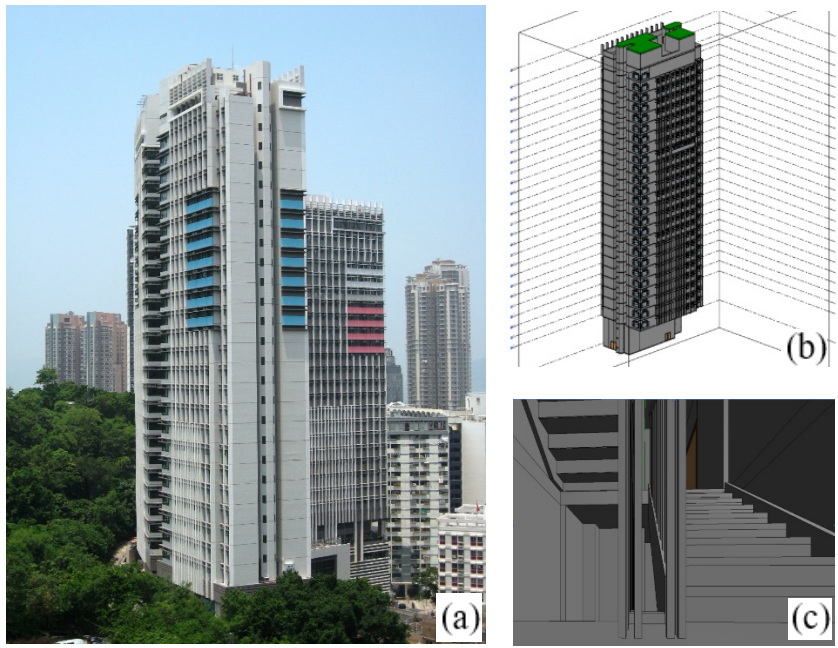

**Fig. 16.** The case study of student residential building at HKU. (a) the real photograph of the building; (b) the external view of the collected BIM model; (c) the rescue stairway to be checked in the building model.

As shown in Fig. 17, an example is given to demonstrate how the proposed method flexibly transforms design requirements into structured queries. The requirement states that "every access staircase in a firefighting and rescue stairway should be provided with landings at the top and bottom of each flight with a minimum dimension of not less than the width of the flight". Nonetheless, the proposed approach cannot directly interpret this because (a) there is disparate naming between the regulation and the design model; and (b) the implications in the regulation require manual understanding. For instance, "dimension" implies the length and width of the landings. Hence, based on the applied terminology and information of BIM models, a translated NLQ consisting of four sentences was formulated: "The stairs in staircase space that have landing slabs (S1). Slabs have length or width less than the actual run width of stair flight (S2). The stair flights that are part of stair (S3). The slabs that have relative height that equals to the relative base height or relative top height of the stair flight (S4)". S1 mentions that the *IfcStair* instances contained in *IfcSpace* instances with a long name of "Staircase", since it was discovered in the residential BIM model that the "access staircase in a firefighting and rescue stairway" was annotated in this manner. Additionally, the landing of a stair in the building model is represented as *IfcSlab* instances with a type enumeration of *LANDING* and being compositions of *IfcStair* instances. Thus, S3 adds a condition that the *IfcStairFlight* instances should be part of the *IfcStair*. S2 demonstrates the quantitative comparison between the slab's dimension and the flight's width. S4 translates the requirement "landings at the top and bottom of each flight" into the condition that landing slabs whose *IfcProperty* "RelativeHeight" equals the *IfcProperty* "RelativeTopHeight" or *IfcProperty* "RelativeBaseHeight" of the *IfcStairFlight* instances. Consequently, taking advantage of the coreference resolution, the interpretation results of the four sentences were merged into one query graph to find out the landings that violate the dimension restrictions.

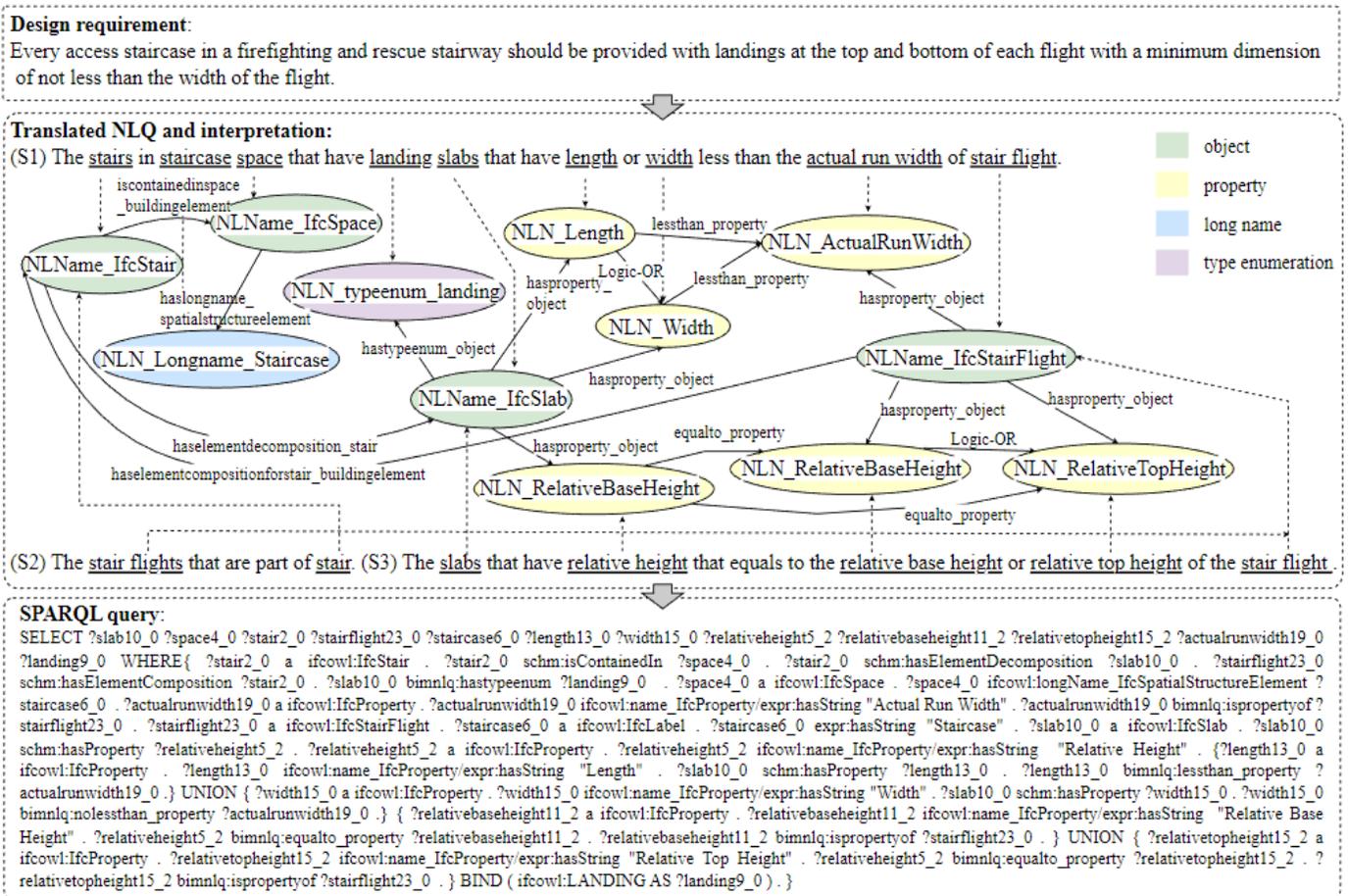

Fig. 17. The case demonstration showing how the proposed method transforms design requirements into standard queries to check design flaws in graph-based BIM models. S1, S2, S3 and S4 represent the four sentences of the translated NLQ.

As a result, all the four design requirements (excluding the geometric calculation parts which are beyond the scope of this study) of the rescue staircase were translated into NLQs successfully. It took around one minute for the formulated programs to interpret multiple-sentence queries and correctly outputted SPARQL queries. These queries were executed upon the residential BIM model, and it turned out to be free of errors because the building design had already been verified previously. This shows the effectiveness of the proposed NL-based method in supporting end-users to effortlessly make complex BIM-oriented queries for information needs in their professional workflows. In contrast, manual formulation of lengthy and logically complete standard queries (see Fig. 17) could be challenging and error-prone for both experts and non-experts in IT. The remainder of the tested regulations, NLQs, and the resulting SPARQL queries can be found in Appendix A.

# 7. Discussion

## 7.1 Comparative analysis

In this section, the SOTA NL-based methods [11,13–15,71] targeted at querying BIM models are compared with the proposed MOP-SP approach in a qualitative way, which compares the claimed functionalities and scopes of different methods. The following aspects are considered:

- Attribute constraints: to what extent can BIM objects be filtered by their attributes?
- Relational constraints: can objects be retrieved by semantic relationships between contextual objects in BIM models?
- Reasoning functions: are reasoning functions supported to retrieve BIM models?
- Multi-constraints expression: can multiple constraints be logically combined in one NLQ?

- Code generation: does the method explicitly illustrate how to generate executable codes from NLs?

The results of the comparison are summarized in Table 11. First, our approach allows retrieving objects with different attributes (e.g., material, type, property), value restrictions (Boolean, quantitative, and literal) and comparison (e.g., attribute A > attribute B), while the SOTA methods can only extract limited attribute constraints that are encoded concepts in IFD, but miss model-specific concepts as well as literal and Boolean value restrictions. Second, our approach models 11 categories of semantic relationships between objects in ontology and provides an efficient relation extraction solution. In comparison, BIMASR [15] only presents the containment relationship between wall and building story as an attribute, and it also lacks a method to classify the intended relationship among several candidates. Third, our approach applies SPIN inference rules for ontology-based reasoning in BIM models, while none of the existing methods involves reasoning functions in NL-based BIM query. Fourth, our approach has logical operation and coreference resolution mechanisms for performing multiple-constraints queries. In contrast, the SOTA methods have limited capabilities to process logical combinations of different constraints in queries. Last, only our method demonstrates how to generate open IFC-compliant SPARQL queries, which suggests better adaptability in various scenarios and data environments.

Table 11. Comparison of the proposed approach with the existing NL-based BIM data retrieval methods.

| Item | Intelli-BIM [11] | BIMASR [15] | iBISDS [12,13] | BIH-Tree [44] | MOP-SP (Ours) |
|---|---|---|---|---|---|
| Attribute constraints | Partial attributes and values predefined in IFD. | Not supported. | Not supported. | Partial attributes and values predefined in IFD. | Complete attributes, data value constraints and comparison. |
| Relational constraints | Not modeled. | Modeled containment as an attribute. | Not modeled. | Modeled containment in BIH-Tree. | Modeled 11 relationships in ontology. |
| Reasoning functions | Nor supported. | Not supported. | Not supported. | Not supported. | SPIN inference |
| Multiple-constraints expression | Logical connection between objects or values in one constraint. | Not supported. | Not supported. | Support conjunction of attributes. | Logical connection between objects or constraints; multiple-sentence queries. |
| Code generation | Not presented. | Not presented. | Not presented. | Not presented. | Presented. |

In sum, our approach allows to use NL text to retrieve BIM models with hybrid user-specified constraint conditions. In comparison to MVD and query language-based approaches [5,9,16,35–37], which require extensive knowledge of data schema and IT skills, an NLI with the proposed semantic parser may be a better approach for building information inquiry (e.g., select objects by attribute) in construction projects, particularly in siteworks where software is difficult to operate.

**7.2 Limitations**

While the research's achievements are promising, several limitations should also be noted.

(a) In NER, the proposed method cannot deal with the ambiguous names that can refer to different entities or literal data values. In such conditions, the proposed method cannot acquire correct results like standard query languages.

(b) Our approach relies heavily on DP. However, when a single query sentence involves more than five variables and irregular morphology of NEs, the resulting DP graphs are erroneous. Hence, it is recommended to use multiple-sentence queries to retrieve BIM models.

(c) The proposed method cannot extract logical relationships between different objects with their own constraints (e.g., "Space has Attribute A or Wall has Attribute B), and conjunct/disjunct value restrictions of an attribute (e.g., "Walls have a height of more than 1 meter and less than 3 meters").

(d) The proposed method requires the input query to explicitly mention all variables because it cannot recognize implicit terms like pronouns (e.g., "it") and cannot extract relationships between variables in the absence of the intermediate variables.

(e) The geometric and spatial information of BIM objects are not yet ready for retrieval. In addition, mathematical calculations and counting functions are not supported yet.

# 8. Conclusion

The natural language-based query approach provides an efficient way to retrieve partial subsets of BIM models in building projects. However, current methods cannot handle natural language queries containing multiple constraints from attribute restrictions and relationships between contextual objects. To tackle the challenge, this paper presents a novel model-based ontology population and semantic parsing (MOP-SP) approach to interpret multiple-constraint NLQs into executable codes for retrieving BIM models. An IFC Natural Language Expression ontology was created to supplement the ifcOWL ontology with NL expressions of IFC concepts and abstracted semantic relationships, and was then populated from BIM models to assimilate model-specific entities. Based on the MOP results, the entities and different forms of constraints in NLQs were progressively parsed into standard SPARQL queries, which could be executed to obtain the desired partial BIM models.

The performance of MOP-SP was evaluated based on 225 collected NLQs. The overall accuracy of the query results is 91.11%. The errors mainly stem from name ambiguity, value restriction, and dependency parsing. Furthermore, a case study was conducted about the design-checking of BIM models against fire safety regulations in Hong Kong. The selected regulations, consisting of complex conditions, can be effectively handled by an NLI with the developed semantic parser to identify design flaws.

The contributions of this research are acknowledged in four aspects.

(a) This paper presents a novel semantic parsing approach that first shows how to automatically convert multi-constraint NL-based queries into explicit IFC-compliant structured queries.
(b) Four types of essential attribute constraints (material, type, property, and quantity) can be used in NLQs to retrieve BIM models with value restrictions and comparison conditions.
(c) Relational constraints can be utilized in NLQs to retrieve BIM objects based on the association with contextual objects.
(d) Different constraints can be logically connected within an NLQ sentence or across multiple sentences, allowing NLs to query BIM models with significantly increased granularity.

The formalized semantic parser can be installed in a variety of IFC-authoring BIM software in the construction industry to allow end-users to retrieve BIM models more efficiently. Moreover, the use of the semantic web and unsupervised text parsing mechanisms ensures the scalability of the proposed method in dealing with diverse project-specific information entities and semantics.

Future research will concentrate on overcoming several limitations of this method, such as the name ambiguity issue. Furthermore, the application scenarios of the semantic parser and NLIs in the whole lifecycle of building projects should be further investigated through more case studies.

# Appendix A

The IFC models, instance RDF files, developed INLE ontology, the NLQ dataset and the case study data used in this research can be accessed via the link: https://github.com/MengtianYin/BIM-NLQI.